\documentclass[sigplan,screen,nonacm]{acmart}
\AtBeginDocument{%
  }

\setcopyright{none} 
\usepackage{subfigure}
\usepackage{algorithm}
\usepackage{algorithmic}
\usepackage{listings}

\begin{document}
\settopmatter{printacmref=false}

\title{EnergonAI: An Inference System for 10-100 Billion Parameter Transformer Models}

\author{Jiangsu Du$^{a,b}$, Ziming Liu$^a$, Jiarui Fang$^a$, Shenggui Li$^a$, Yongbin Li$^a$, Yutong Lu$^b$, Yang You$^{*,c}$}
\affiliation{
 \institution{HPC-AI Technology Inc.$^a$ \hspace{0.3em} Sun Yat-sen University$^b$ \hspace{0.3em} National University of Singapore$^c$}
}

\thanks{The work was done when Mr. Du was an intern of HPC-AI Technology Inc.}
\thanks{* Corresponding Author (youy@comp.nus.edu.sg). Yang You is a faculty member at NUS; this work was done at HPC-AI Technology Inc.}

\begin{abstract}
Large transformer models display promising performance on a wide range of natural language processing (NLP) tasks.
Although the AI community has expanded the model scale to the trillion parameter level, the practical deployment of 10-100 billion parameter models is still uncertain due to the latency, throughput, and memory constraints.

In this paper, we proposed EnergonAI to solve the challenges of the efficient deployment of 10-100 billion parameter transformer models on single- or multi-GPU systems.
EnergonAI adopts a hierarchy-controller system architecture to coordinate multiple devices and efficiently support different parallel patterns.
It delegates the execution of sub-models to multiple workers in the single-controller style and applies tensor parallelism and pipeline parallelism among the workers in a multi-controller style.
Upon the novel architecture, we propose three techniques, i.e. non-blocking pipeline parallelism, distributed redundant computation elimination, and peer memory pooling.
EnergonAI enables the users to program complex parallel code the same as a serial one.
Compared with the FasterTransformer, we have proven that EnergonAI has superior performance on latency and throughput.
In our experiments, EnergonAI can achieve 37$\%$ latency reduction in tensor parallelism, 10$\%$ scalability improvement in pipeline parallelism, and it improves the model scale inferred on a single GPU by using a larger heterogeneous memory space at cost of limited performance reduction.


\end{abstract}


\maketitle

\section{Introduction}




Large transformer models~\cite{brown2020language, smith2022using} trained on massive text collections have proven promising capabilities for natural language processing (NLP) tasks.
Although the pre-trained usage mode and efforts on open-sourced parameters~\cite{zhang2022OPT, bigsciencebloom} significantly reduce hardware requirements in using large transformer models, the inference of large transformer models still blocks researchers and engineers from studying and deploying downstream tasks.
To handle the single-instance multiple-device inference paradigm, existing solutions are heavily affected by the single-device inference system and the distributed training system, leaving room for both performance and programmability.

Large transformer models are astonishing in size, leading to unafforadable computation and memory demands.
Therefore, different from traditional small-model inference systems, additional memory management, and communication techniques have to be applied to extend the large-model inference systems on multi-GPUs.
However, direct immigration of these techniques from large model training system is not suitable.
Although the inference is the feed forward process of the training, the model training mainly targets at high throughput, while the model inference targets both throughput and latency.
Therefore, there exists a huge gap in optimization strategies between them.
More importantly, the poor programmability is always criticized that existing inference frameworks are implemented highly coupled using C++/CUDA and can be hardly customized for new models, not to mention the distributed inference with complicated memory management and communication logic.

An efficient transformer inference system is still the missing piece in the large model blueprint.
This is because the large model inference is an unique problem different from small model inference and big model training.
First, small model inference frameworks are highly coupled with customized C++/CUDA kernel implementations~\cite{fast2022former, bytedanceeff}. 
It is non-trivial to customize them for new models.
Further, we observed that these traditional optimization methods, such as kernel fusion and customized parallel kernels, have little effect on the large models.
Second, existing large transformer model inference systems\cite{rasley2020deepspeed, fast2022former, hanetal2022bminf} take advantage of the parallel approaches from training systems.
Although both training and inference embrace a similar computation pattern, the model training mainly targets high throughput, but the model inference targets both throughput and latency.
Therefore, directly copying the training strategies is not the optimal solution.
To fill the missing piece, we believe that the unique challenges of large model inference must be addressed.

We identify the most critical problem in the context of large-model inference is how to break the memory wall which limits the available model scale and sometimes lower the computing efficiency.
A line of work~\cite{rasley2020deepspeed, hanetal2022bminf} proposed to offload parameters to the host memory.
However, the extremely low PCI-e bandwidth also damages the performance.
An line of work break memory wall by extending inference tasks to multiple GPUs.
The most common approach is to parallelize a model based on the single program multiple data (SPMD) model~\cite{rasley2020deepspeed, fast2022former}.
The SPMD model takes the multi-controller architecture that each worker executes the similar computation and all workers communicate through collective primitives.
However, the SPMD model is not efficient for the distributed inference to control multiple devices with irregular workloads.
For example, it is difficult for the SPMD model to efficiently implement the pipeline parallelism.
Since different workers perform different stages of a model in pipeline parallelism, it requires new coordination primitives beyond collective communications.
Also, to keep the consistence of consecutive devices, these coordination primitives has to be in the blocking style~\cite{fast2022former} and result in more communication overheads.

In this paper, we describe our distributed system, EnergonAI, which targets at inferring 10-100 billion parameter level transformer models.
EnergonAI considers the large transformer inference from the bottom-up and takes both performance and programmability into account.
In summary, our contributions are as follows:
\begin{itemize}
    \item We make an in-depth empirical performance profiling and parallel behavior analysis of the large transformer inference, which provides insights on system design.
    \item We take the best of both single-controller and multi-controller parallel architectures, and propose a novel hierarchy-controller system architecture.
    It includes a distributed runtime and a centralized engine to smartly combine tensor parallelism and pipeline parallelism for the large transformer model inference. Also, the hierarchy-controller system architecture guarantees the distributed inference the same behavior as the serial inference.
    \item We propose three techniques, i.e. non-blocking pipeline parallelism (NBPP), distributed redundant computation elimination (DRCE), and peer memory pooling (PMEP) to further improve the metrics on latency, throughput, and resolve the memory wall problem.
\end{itemize}

Here some performance results of EnergonAI are displayed.
Compared with the single-device inference, EnergonAI can achieve up to 88.2$\%$ latency reduction using tensor parallelism on 8 GPUs, and 3.82$\times$ throughput growth on 4 GPUs using pipeline parallelism.
When the number of GPUs is 4, EnergonAI has about 37$\%$ latency reduction compared with the competitor, FasterTransformer~\cite{fast2022former}, in tensor parallelism, and the scalability of EnergonAI is 10$\%$ better than that of FasterTransformer in pipeline parallelism.
Also, a case study shows that the model scale supported on a single GPU device can be doubled with only 4$\%$ latency reduction.

\section{Background}

\subsection{Neural Language Model Pretraining}
Pretrained language models (PLMs) have become indispensable artifacts for the NLP field.
PLMs can learn the general language knowledge from the large corpus and exploit this general knowledge for specific language tasks.
Early usage of PLMs is the word embedding, such as Word2Vec~\cite{mikolov2013efficient} and GloVe~\cite{pennington2014glove}.
Word embedding table transfers natural words into digital representations.
The pretrained word embedding technique uses neural network models to learn word associations from a large corpus of text and produce the word embedding table, which can largely improve downstream tasks.
Later works~\cite{devlin2018bert, yang2019xlnet, radford2019language} advanced research in this area by further learning from a larger corpus and taking transformer-based model structures.
These PLMs require a finetuning stage and only require a small corpus for specific tasks.
Next, PLMs continuously become larger and start exploring the few-shot or even zero-shot learning~\cite{brown2020language, smith2022using}, which only requires a few or even no samples for specific tasks.

\subsection{Transformer Models and Parallel Patterns}

\begin{figure*}[!t]
    \centering
    \includegraphics[width=0.85\textwidth]{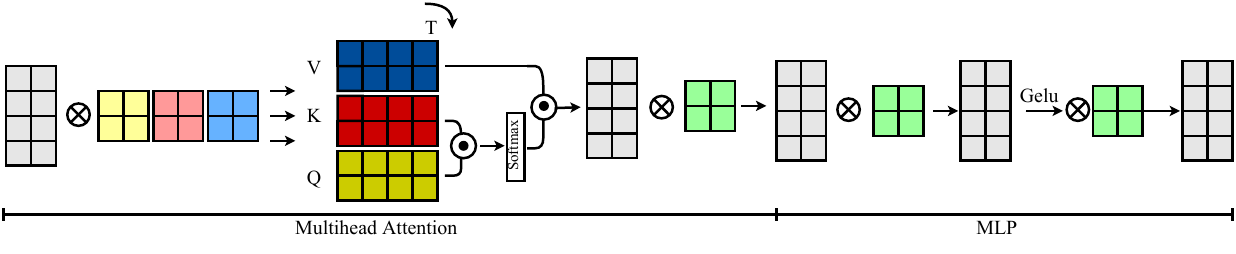}
    \caption{The illustration of a single transformer layer.}
    \label{fig:transformer}
    \vspace{-10pt}
\end{figure*}

Transformer-based models are a series of models mainly follow the structure of Transformer~\cite{vaswani2017attention}, and here we call them Transformer models in brief.
Original Transformer was designed for the machine translation task, employing two major architectures, encoder and decoder. 
While recent transformer models, e.g. BERT and GPT~\cite{brown2020language}, only stack one of the encoder and decoder as the component.
In inference process, the encoder and decoder differ slightly.
Decoder has an extra masking structure (casual mask) and it does not have an impact on parallel behavior.
Besides the encoder or decoder, a complete transformer model has an embedding layer that is responsible for converting the natural language sentence into the digital representation.

There are three potential paradigms in making transformer inference parallel, i.e. data parallelism, tensor parallelism, and pipeline parallelism.
For data parallelism, it replicates a model multiple times, and can only benefit throughput.
For tensor parallelism, it partitions a transformer layer horizontally into different devices.
Although it has a high demand on the performance of the interconnection, it is potential to simultaneously optimize latency, throughput, and memory footprint of model inferences.
For pipeline parallelism, it partitions a model by layers, and it can optimize throughput and memory footprint, other than latency.
Figure~\ref{fig:transformer} illustrates the structure of a single transformer layer. 
Basically, a transformer layer is made up of a multi-head attention module and an mlp module.
Grey blocks represent the intermediate results, i.e. activation, blocks in light color represent parameters, and blocks in dark color represent query, key, and value of the attention module.
The multi-head attention module includes a pair of linear layers and a multi-head attention structure.
The mlp module only includes a pair of linear layers.

\section{Design Motivation}

The design choices of systems are often driven by the properties of the underlying target hardware and the upper workloads.
Compared with small model inference and distributed training, 
here we introduce two main differences in the large model inference, which largely motivates our design of the EnergonAI.

\subsection{Performance v.s. Programmability}
\label{sec:pp_motivation}

\begin{figure}[!t]
    \centering
    \includegraphics[width=0.9\columnwidth]{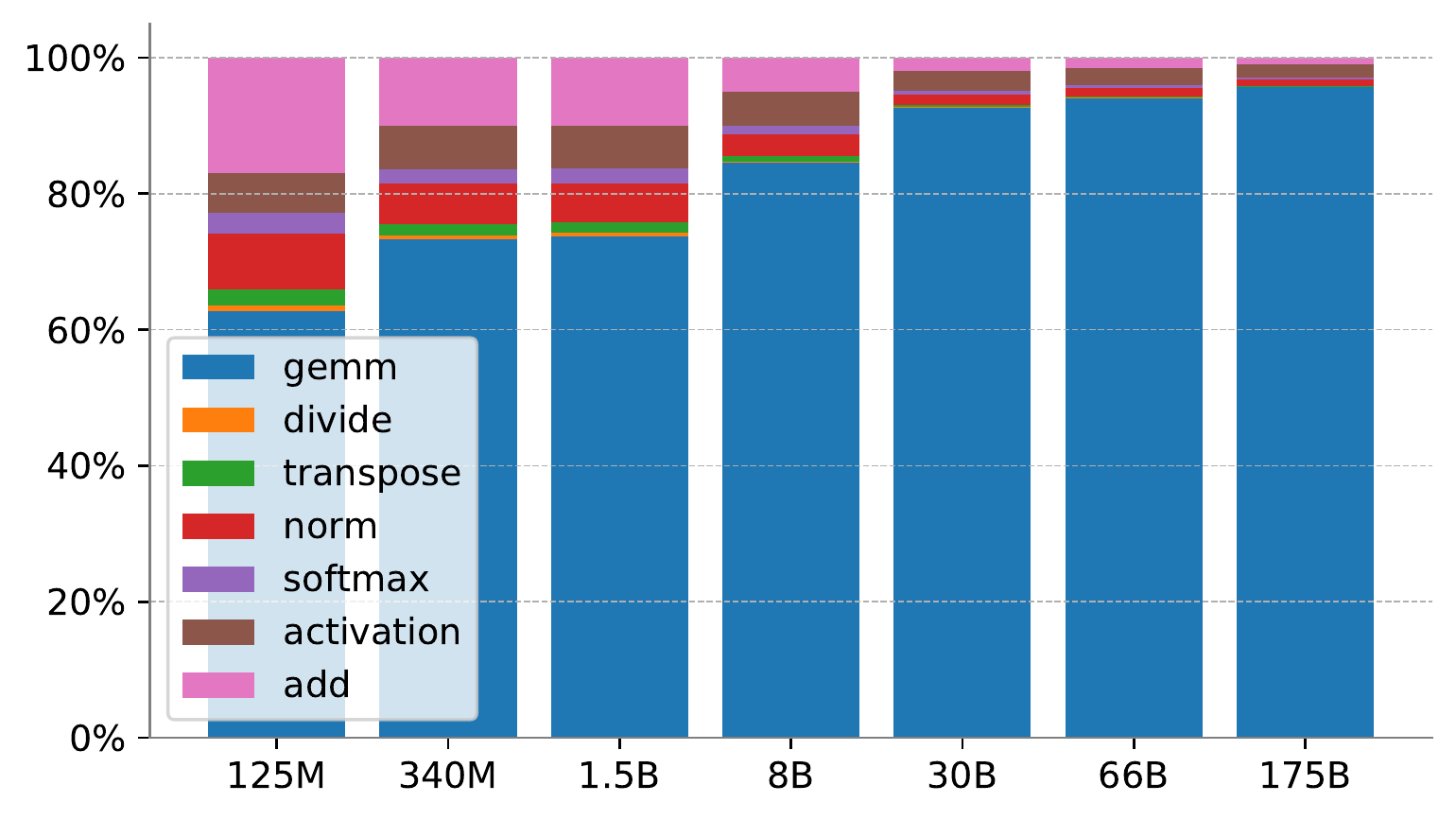}
    \caption{Normalized kernel execution time distribution of different model scale on A100 GPU. Here we select GPT model series with batch size 32, sequence length 64, and precision FP16.}
    \label{fig:kernels}
    \vspace{-10pt}
\end{figure}

To pursue the optimal performance, inference frameworks~\cite{fast2022former, jiarui2020turbo} of small transformer models, are implemented highly coupled directly using C++/CUDA.
Although it can solve the performance-related problem, the poor programmability is always criticized.
Since there lacks modularity and abstraction design, the memory management and computation are mixed together, making it hard for identifying a module, such as a linear layer.
For example, there are complex conditional statements in FasterTransformer~\cite{fast2022former}, such as selecting the precision.
Thus, it is difficult for users to customize the memory access and computation patterns when deploying a new model.

Moving onto large models, users should even take communication logic into account, imposing the burden on knowing HPC-related knowledge.
In the training system, a neural network is typically abstracted into different layers.
Dynamic languages and binding techniques are adopted for encapsulating memory management and computation together as a layer.
In this way, in our EnergonAI, we tend to keep the concept of layer and encapsulate computation, memory management, and communication logic together as the distributed layers for improving the usability.

For the small model, the kernel fusion technique fuses two or more independent kernels into a single kernel for improving the performance, especially for these memory-intensive operations.
The usage of fused kernels can invade into different layers, breaking the abstraction design.
Luckily, as model sizes surge by more than two orders of magnitude, the computation feature of large transformer models can better support the design.
Figure~\ref{fig:kernels} presents the normalized kernel execution time distribution of GPT models in different sizes on NVIDIA A100 GPU.
Since a single A100 GPU is not capable of accommodating GPT-66B and GPT-175B, here we employ their configurations and collect execution time in a single transformer layer.
We select 32 as the batch size, 64 as the sequence length, and FP16 as the precision in these experiments, which can saturate the GPU and reach the largest throughput.
Figure~\ref{fig:kernels} shows that the time ratio of general matrix multiplication (GEMM) kernels increases from about 62$\%$ to about 96$\%$ as the model size grows from 125 million to 175 billion.
In other words, these compute-intensive kernels become more dominant in large model inference.
Consequently, the motivation of kernel fusion is largely weakened.

\subsection{Control FLow Architecture}

As the model size has largely exceeded that a single GPU's memory can accommodate, large models require the collaboration of multiple devices or even multiple nodes, making the large model inference system a distributed system.
Here, we discuss how different control flows shape a distributed ML system and their pros and cons of them.

Distributed ML systems mainly have two control flow architectures~\cite{barham2022pathways}, i.e. the multi-controller architecture and the single-controller architecture.
Multi-controller architecture is usually adopted in the distributed model training for its simplicity.
This architecture asks each device to load a replica of the user's code and decide what it performs based on the process rank number.
Each device executes a similar workflow which involves computation and collective communications.
MPI~\cite{clarke1994mpi} in HPC applications, ColossalAI~\cite{bian2021colossal} and Deepspeed~\cite{rasley2020deepspeed} for large model training are classical examples of such architecture.
However, this architecture is poor in handing pipeline parallelism and irregular models, because they are not symmetrical and require new coordination primitives beyond collective communications.
And, pipeline parallelism has already been an important parallel pattern in situations where the interconnection network is poor in bandwidth.
On the contrary, the single-controller architecture generally requires a centralized manager that is responsible for dispatching tasks across multiple devices.
In comparison, this architecture has better flexibility and is easy to support centralized resource management and scheduling.
The defect is that a general single-controller architecture can be extremely complicated to implement.

Unfortunately, 
the above two control flow architectures can poorly match the requirements of distributed inference system.
The multi-controller architecture can efficiently manage communications of tensor parallelism and achieve good performance, while it is not suitable for pipeline parallelism.
The single-controller architecture is straightforward to implement the pipeline parallelism given the centralized manager has a global view.
However, it is difficult to conduct fine-granularity control of workers' behavior, e.g. communications, through the master node, hardly achieving good efficiency.
In this way, there needs a trade-off between multi-controller and single-controller architectures.

\section{EnergonAI}

\begin{figure}[!t]
    \centering
    \includegraphics[width=0.95\columnwidth]{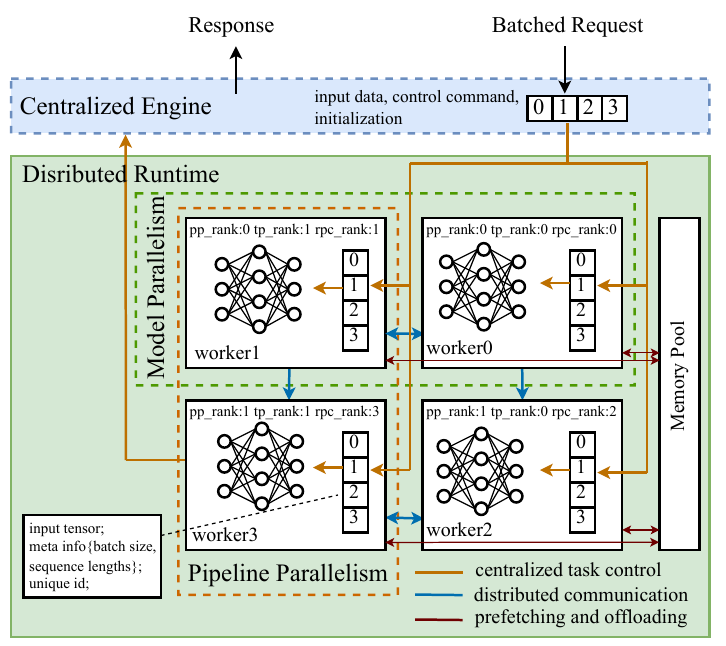}
    \caption{The overview of EnergonAI.}
    \label{fig:overview}
    \vspace{-10pt}
\end{figure}

Figure~\ref{fig:overview} shows the overview of EnergonAI.
It is abstracted into a two-layer architecture, i.e. the centralized engine and the distributed runtime.
Distributed runtime is built upon Pytorch~\cite{paszke2019pytorch}.
It includes a global communication context and some distributed operations.
Centralized engine is an abstraction layer newly introduced for managing distributed multi-device inference.
It is responsible for the centralized task publish and schedule using the remote procedure call (RPC).
Consequently, the centralized engine implements a RPC communication context.
With a careful encapsulation design, the engine layer makes the distributed multi-device inference of EnergonAI the same usage behavior as the single-device inference.
Besides the basic architecture, there are three techniques for improving the computation efficiency and relieving memory wall problem, i.e. non-blocking pipeline parallelism (NBPP), distributed redundant computation elimination (DRCE) and peer memory pooling (PMEP).

\subsection{Hierarchy-controller System Architecture}

To make multiple devices collaborate better, we take the best of both multi-controller architecture and single-controller architecture, proposing the hierarchy-controller system architecture for EnergonAI.
For the distributed runtime, operations are programmed in the style of SPMD model, knowing what data and which device it should communicate, like in the multi-controller architecture.
For the newly abstracted engine, it takes the centralized management that it controls the initialization and task launch on workers from a global view, like in the single-controller architecture.
Basically, there are two sets of communication contexts, i.e. the global communication context and the RPC communication context.

\subsubsection{Distributed Runtime}

Constructing the distributed runtime follows the idea of the SPMD model.
To manage the communication meta information, e.g. world size and rank number, a global communication context is designed.
Based on the global communication context, we implement distributed operations for tensor parallelism.
These distributed operations include the computation and communication logic that perform the related computation and use communications to remove data dependencies.
For each device, it knows what data it should compute, what data it should communicate, and which device it should communicate to based on the global communication context.

\subsubsection{Centralized Engine}

The centralized engine manages and controls workers by RPC.
It is mainly responsible for two separate stages, i.e. runtime initialization and execution launch.
To support the engine, an RPC communication context is required to initialize at the beginning.
For the runtime initialization, it delegates sub-models to workers, initializes the related part of the model, loads parameters into memory, etc.
For the execution launch, it sends input and control information to related workers.
With the centralized management of the engine, it is more efficient to perform these irregular workloads, which will be explained in the part of Section~\ref{sec:tech1}.

\subsubsection{1-D Tensor Parallelism}
\begin{figure}[!t]
    \centering
    \includegraphics[width=0.7\columnwidth]{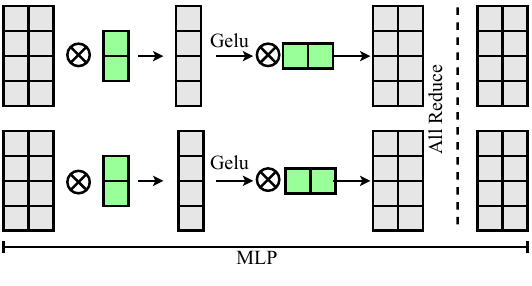}
    \caption{The illustration of a single mlp module in 1-D tensor parallelism.}
    \label{fig:transformer1d}
    \vspace{-10pt}
\end{figure}

To distribute the transformer model in tensor parallelism, we follow the partitioning of the 1D strategy in Megatron-LM~\cite{shoeybi2019megatron}. 
In the strategy, as shown in Figure~\ref{fig:transformer1d}, the linear operation requires extra communications for removing data dependencies when distributing and other operations can be directly applied without extra communications.

Linear operation uses matrix multiplication to transform input features into output features using a parameter matrix.
Since linear operations always appear in pairs in transformer models, here are chances to consider them as a unity and reduce communications.
The mlp module in Figure~\ref{fig:transformer} can display two continuous linear functions in matrix form.
Grey matrices represent intermediate features and green matrices are parameters.
To distribute two consecutive linear operations into two devices, as in Figure~\ref{fig:transformer1d}, each device has the same input and the 1-D strategy splits the parameter matrix of the first linear operation in column.
Then, the 1-D strategy splits the parameter matrix of the second linear operation in row.
Finally, the output matrices in different devices should be accumulated by communications.
Consequently, there only requires a single synchronization point every two linear operations.
It is also feasible for the self-attention module, in which the output of the first linear layer in each device is the input of corresponding attention heads.
The linear operation is implemented in two modes, i.e. column and row, and each device can know what to perform based on the global communication context like the style of MPI.

As shown in Figure~\ref{fig:overview},
the engine performs in a centralized style and it sends model architecture implemented using the distributed runtime to each worker for initialization.
Then these workers will initialize the related part based on the global communication context.
When new input comes, the engine will send a command to tell each worker the related information about the input.
After receiving the command from the engine, each worker knows how to execute the inference of the batch.
Then the concrete execution for a single batch is performed in the style of SPMD model.

\subsection{Technique-1: Non-blocking Pipeline Parallelism}
\label{sec:tech1}

\begin{figure}[!t]
    \centering
    \includegraphics[width=0.95\columnwidth]{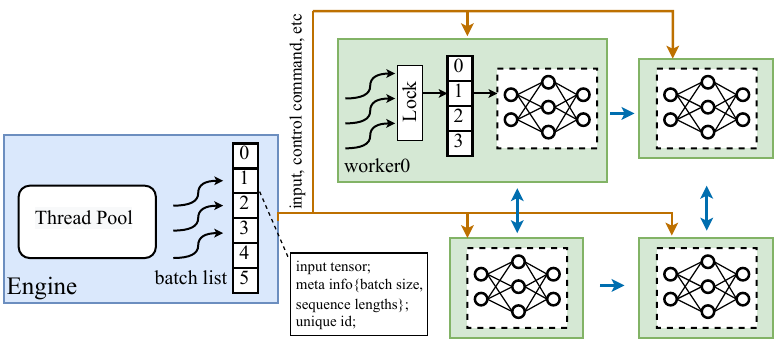}
    \caption{Engine Management of Tensor Parallelism and Pipeline Parallelism.}
    \label{fig:engine}
    \vspace{-10pt}
\end{figure}

For pipeline parallelism, it partitions a model vertically.
For the transformer model, it is generally partitioned by transformer layers.
In other words, different transformer layers will be allocated to different devices.
Similar to tensor parallelism, pipeline parallelism is also performed under the hierarchy-controller architecture that the engine sends commands to each worker and each worker knows what data and which device it should communicate.
Here we design our pipeline parallelism in a non-blocking style pursuing an excellent performance.

At first, the engine works in the non-blocking style that it launches a task to workers and returns immediately.
For pipeline parallelism, multiple inputs will be processed simultaneously and locate in different pipeline stages.
In this way, it is necessary for the engine to launch multiple tasks in the non-blocking style.
Second, workers should also work in the non-blocking style that it takes the asynchronous communications for removing data dependencies between two consecutive devices.

However, it is non-trivial to achieve the two-stage non-blocking execution.
For example, if multiple threads are launched by RPC for performing different samples in a worker and threads compete for the lock, winning threads in different workers might relate to different samples.
In this way, the corresponding relationship of the input and the result is wrong.
Moreover, if the batch size and padding size are variable in different batches, the system can be in a dead lock since the intermediate results are different in size.
In this way, the naive implementation of pipeline parallelism generally adopts the synchronize communication to keep the consistence between consecutive devices. 

To enable the 2-stage non-blocking execution, we add two designs in the engine.
First, as shown in Figure~\ref{fig:engine}, we use a thread pool in the engine that fetches batches from the batch list.
For a single thread in the engine, it will launch an inference task to workers with input tensor and some meta information.
For each worker, a thread is correspondingly created and receives the command from the engine.
In this way, the non-blocking execution of the engine is achieved.

Second, to make sure that different workers process the same input in a single inference, we customize a distributed consistency queue for keeping the execution order of multithreading.
In the engine and all workers, we keep a loop data structure that increments unidirectionally.
The engine will get an updated value from the loop data structure as the unique key and add it to the command.
After receiving commands from the engine, the worker will add the batch into the local queue and use the remote unique value as a key.
In this way, for the thread that acquires the lock, it will not execute the batch it originally receives, while it takes the local unique key from the local loop data structure, find the batch uses the local key, and executes the batch.
Thus, batches are processed in the order of arrival.
Since the engine and all workers follow the loop data structure, all workers can keep consistence and execute the same input in a single inference.
With both thread-pool and distributed consistency queue designs, it is feasible to use the asynchronous communication.
Consequently, each worker will constantly and independently perform computation without waiting communication.
Moreover, as shown in Figure~\ref{fig:engine}, it is straightforward to combine both pipeline parallelism and tensor parallelism under the hierarchy-controller architecture.

\subsection{Technique-2: Distributed Redundant Computation Elimination}

\begin{figure}[!t]
    \centering
    \includegraphics[width=0.95\columnwidth]{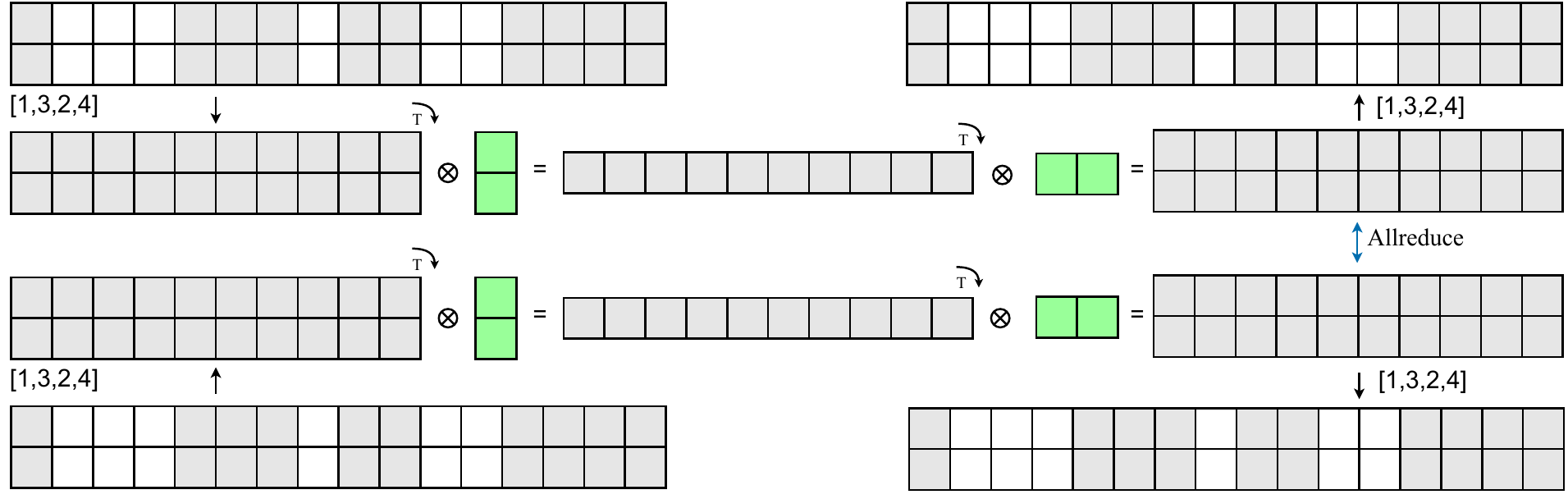}
    \caption{Distributed Redundant Computation Elimination.}
    \label{fig:drce}
    \vspace{-10pt}
\end{figure}

Since transformer models are majorly used for natural language tasks and natural language sentences are variable in length~\cite{du2022handling}, the lengths of sequences in the same batch vary a lot.
Padding them into the same length introduces a big waste of computation resource.
Thus, we propose the distributed redundant computation elimination (DRCE) technique in EnergonAI.
Basically, EnergonAI supports eliminating the redundant computation in the MLP module.

To eliminate the redundant computation in the MLP module, we extend the single-GPU idea from ByteDance~\cite{bytedanceeff} to multi-GPU distributed inference.
As shown in Figure~\ref{fig:drce}, since the vector of each word multiplies with parameters individually in the linear operation, we can accumulate words from different sentences in a batch together.
However, the multi-head attention module still requires the padding area.
Thus, we can remove padding before the multi-head attention module and rebuilds padding after the multi-head attention module, eliminating the redundant computation in all linear layers.

To remove and rebuild padding, we need to tell all workers the information of sequence lengths in a batch in the distributed context.
Benefiting from the centralized management of the engine, it is intuitive to add the information into the command.
Thus, we bind the sequence length information of a batch with the command of the engine and the engine will send it to all workers.
Besides, to support DRCE, here needs the process of removing and rebuilding padding.
Since the shape of input and output keeps unchanged before and after the multi-head attention module, we can directly add the removing and rebuilding process.
To accelerate the data layout switch process, two CUDA kernels that fuse the transpose and pad operations are bound into the system.

\subsection{Technique-3: Peer Memory Pooling}
\begin{figure}[!t]
\centering 
\includegraphics[width=0.45\textwidth]{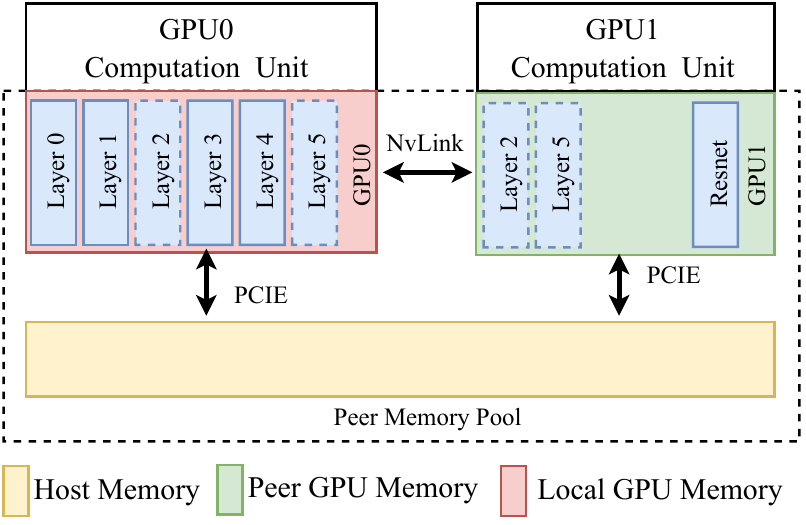} 
\caption{The peer memory pool contains host memory and all GPU memory within the same node. Parameters of large models are distributed into different GPUs (host memory is also the candidate) by layers and parameters of the corresponding layer will be loaded to the computing GPU only when they are about to be used.}  
\label{fig:offload}
\vspace{-10pt}
\end{figure}

Although employing more devices for tensor or pipeline parallelism for the large model inference can alleviate the memory wall problem, it is not cost-effective when large models are not in great demand.
In other words, there are cases that only partial GPUs in a node work for the large model inference facing the memory wall problem, while other GPUs still have considerable memory remains to be used.
Benefiting from the serialized structure of deep learning models and the high-speed interconnection, as in Figure~\ref{fig:offload}, we are inspired to introduce peer memory pooling technique, which treats memory of multiple GPUs within a node as a unity for enabling large model inference.
Moreover, although memory from peer GPUs is our main focus, the host memory and any other memory resources can also be adopted in the pooling technique for enlarging the model size that limited number of GPUs can handle.

Two prerequisites make the peer memory pooling technique feasible.
Here we call the GPU that executes the computation of large model inference as the local GPU and the GPU that only contributes its memory as the peer GPU.
One of the prerequisites is that the extra communication does not cause too much negative effect on peer GPUs.
In theory, NVLINK\cite{Foley2017nvlink} of A100 has 600GB/s bandwidth and HBM of A100 has 1555GB/s bandwidth, which there exists a huge gap.
Here we take ResNet50~\cite{he2015resnet} with TensorRT running on NVIDIA A100 with NVLINK for experiments. 
When we perform a full bandwidth communication between two GPUs, the ResNet50 inference on one of the GPUs has only below $5\%$ performance degradation, and there even has no performance degradation in some cases.

Another prerequisite is that the communication bandwidth is enough for overlapping the communication and computation in the large model inference.
For instance, the bandwidth between two devices connected with NVLINK reaches 600GB/s and one layer of GPT3-175B contains about $1.812\times10^9$ parameters and takes up 3.375 GB memory if we use half precision.
In this case, the overhead of offloading one layer of the model is only about $5.63\times10^{-3}$ second and communication can be easily overlapped with proper pre-fetching.
In addition, the fact that the bandwidth will not be fully used can enhance the first prerequisite

As shown in Figure~\ref{fig:offload}, the peer memory pooling technique treats all memory in a node as a unity and stores parameters of a large model into the pool.
In the memory pool, parameters of a large model are distributed into memory of different GPUs by layers.
Through monitoring memory space on all GPUs, the memory pool decides which device is available for offloading the layer.
Since the bandwidth between CPU and GPU is much smaller than that of NVLINK connections, CPU memory is only used when we exhaust all peer GPU memories, causing a severe performance reduction.
Next, our strategy of offloading is that the layers to be offloaded are decided before the inference starts by the global manager. 
Layers to be offloaded are distributed evenly among those to be held on device. 
To achieve communication and computation overlap, an loading process of an off-device layer is launched before the forward-propagation of the previous layer starts and the offloading process is launched immediately after the computation's done.

\begin{figure}[!t] 
\centering 
\includegraphics[width=0.45\textwidth]{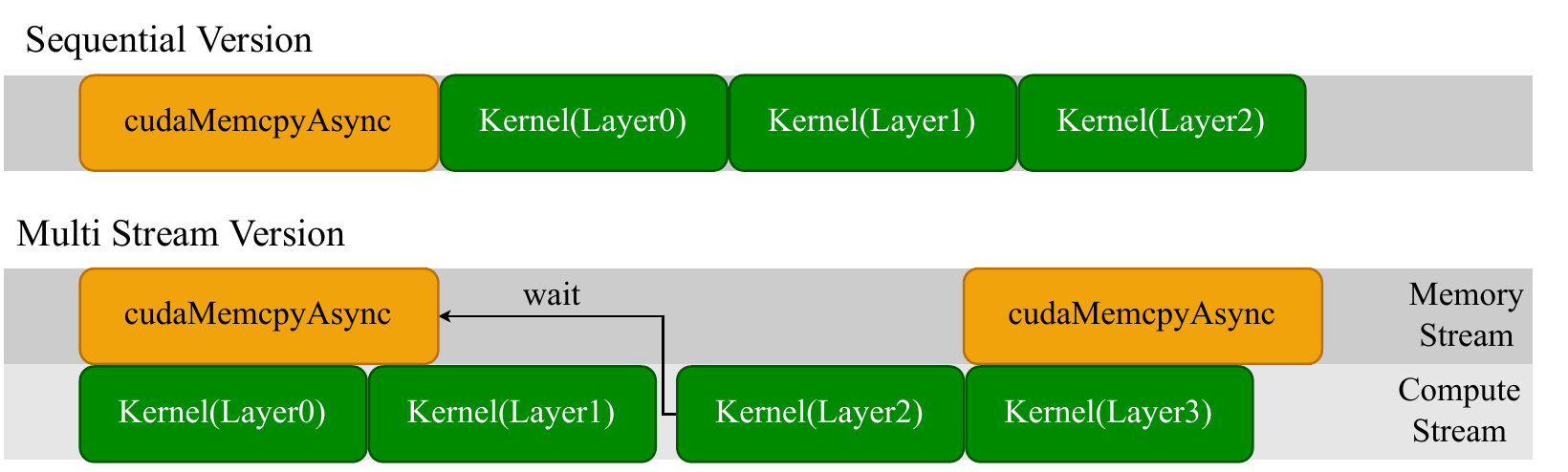} 
\caption{Asynchronous Layer Prefetching.}  
\label{fig:multistream}
\vspace{-10pt}
\end{figure}

To overlap communication with computation, an intuitive strategy is to prefetch a layer before executing asynchronously.
Here we prefetch parameters of a layer a certain number of layers before its execution.
The exact layer number depends on the ratio of communication and computation, leaving a trade-off.
Besides, it is essential to adopt the multi-stream pattern for enabling the real asynchronous execution of layer prefetching.
As in the sequential version of Figure~\ref{fig:multistream}, the blocking or unblocking API can only control whether the CPU launches commands and returns immediately.
To achieve the asynchronous execution, we need to arrange the Memcpy kernels and compute kernels into different CUDA streams.
As in the multi-stream version of Figure~\ref{fig:multistream}, we launch the cudaMemcpyAsync kernel for prefetching parameters of layer2 before launching these compute kernels of layer0.
After executing layer0 and layer1, we need to check whether the cudaMemcpyAsync kernel in stream0 has finished, or rather the corresponding parameters of layer2 are in the local GPU.

\section{Evaluation}
\subsection{Experimental Setup}

Experiments are made in two GPU servers with different configurations.
The first server contains 2 AMD EPYC 7742 CPUs and 8 NVIDIA A100-80 GPUs (fully connected with NVLINK) and the second server contains 2 AMD EPYC 7543 32-Core CPUs and 8 NVIDIA A100-80 GPUs (every two GPUs connected with NVLINK).
Since GPT-3~\cite{brown2020language} is one of the most popular architectures in large transformer models, we select its configuration (head number 96 and head size 128) for the transformer layer in our experiments.
For example, here we call a customized model with 12 layers in GPT-3 configuration as 12-layer GPT-3.
All these experiments are made in FP16.

We evaluate EnergonAI in two steps.
First, we show the pure tensor parallelism scalability of EnergonAI on the fully NVLINK connected GPU server.
Second, we evaluate three techniques respectively on the partially NVLINK connected GPU server, employing NVIDIA FasterTransformer as the baseline.

\subsection{Usability}

\begin{figure}[!t] 
\centering 
\includegraphics[width=0.8\columnwidth]{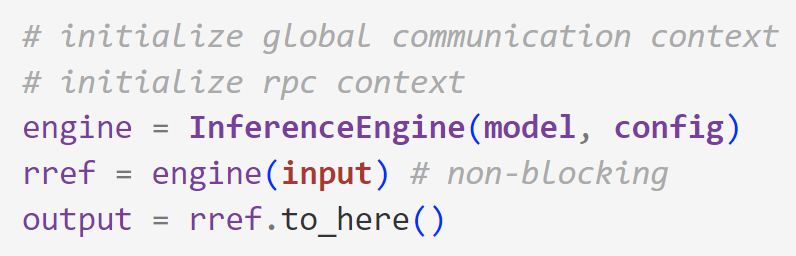} 
\caption{The usage of EnergonAI.}
\label{fig:usability}
\vspace{-10pt}
\end{figure}

Figure~\ref{fig:usability} shows the usage of EnergonAI.
Users need to write the model architecture as in Pytorch.
By contrast, there requires to use distributed operations of our distributed runtime for tensor parallelism and specify the vertical partitioning for pipeline parallelism.
Next, we provide a launch tool for initializing the global communication context and the RPC context.
User can specify the size of tensor parallelism and pipeline parallelism in the launch tool.
Finally, the distributed multi-GPU inference can be used in the same style of the serial single-GPU inference.
After getting the remote reference from the engine, users can fetch the result whenever it is required.

\subsection{Tensor Parallelism Scalability}
\begin{figure}[!t] 
\centering 
\includegraphics[width=0.95\columnwidth]{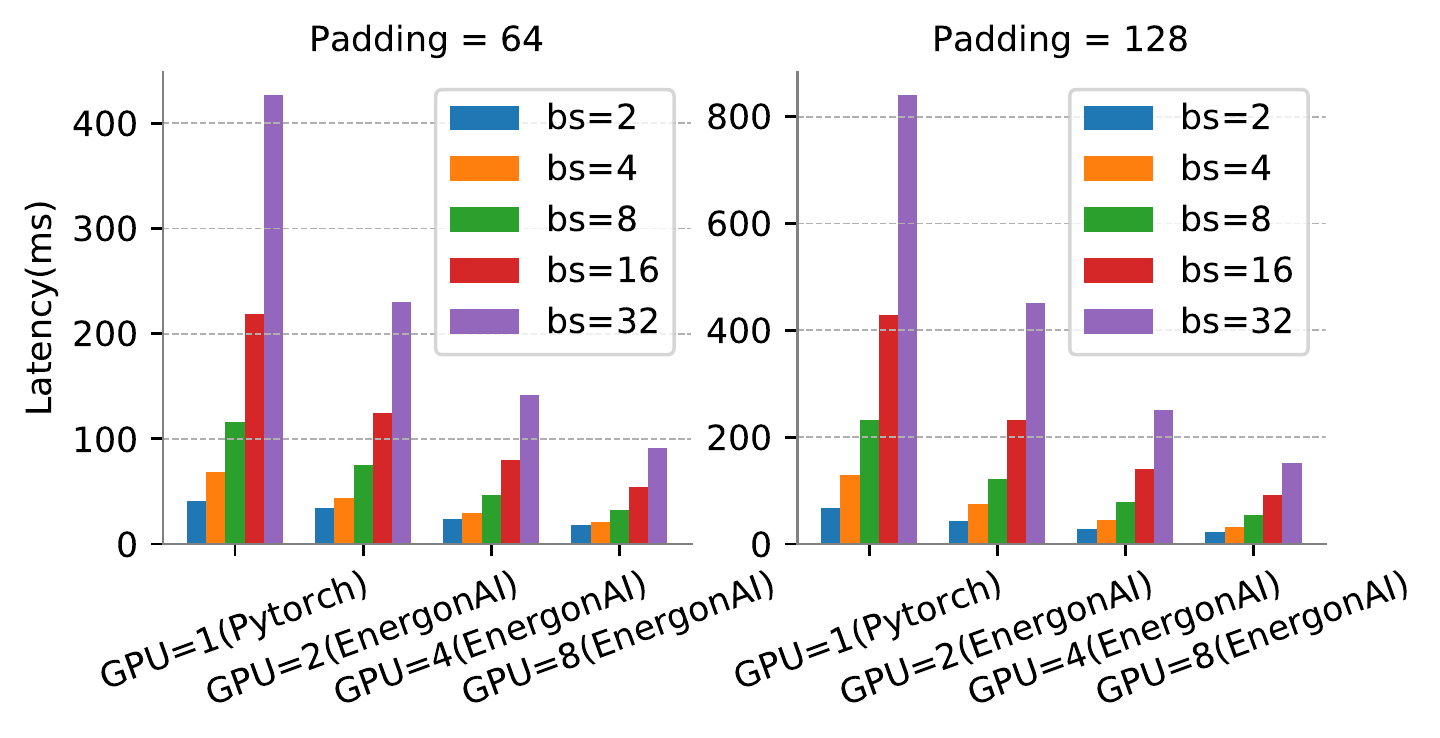} 
\caption{Tensor parallelism performance of EnergonAI on the fully connected GPU server. Pytorch is the baseline.}
\label{fig:basic_speedup}
\vspace{-10pt}
\end{figure}

This section evaluates the pure tensor parallelism of EnergonAI.
To fit the model into a single GPU, we customize a 12-layer model (about 44GB parameters) in GPT-3 configuration.
As shown in Figure~\ref{fig:basic_speedup}, we take Pytorch on a single GPU as the baseline and scale EnergonAI to 2, 4, and 8 GPUs of the fully NVLINK connected server.

As shown in Figure~\ref{fig:basic_speedup}, we can notice that inputs with larger batch size and larger padding size show better speedup in the distributed inference.
For example, when the GPU number increases from 1 to 8, the input with batch size 2 and padding size 64 only achieves 55.8$\%$ latency reduction, while the input with batch size 32 and padding size 128 can achieve 82.0$\%$ latency reduction.
This comes from two potential reasons.
From the view of computation, input with the small batch size and padding size generally cannot saturate computing units in GPUs, and splitting the workload into pieces can further exacerbate the problem, resulting in low scalability.
From the view of communication, although the communication volume is proportional to both batch size and padding size, the communication overhead is relatively larger when the communication volume is small.
This is because there are fixed overheads other than the practical data transfer in communications.
Thus, it is better to infer more samples together when using tensor parallelism.

Even if we use the fully NVLINK connected server,
We find that tensor parallelism is inefficient when scaling to many devices.
This is because each transformer layer requires to communicate and remove dependencies, introducing large amounts of communication overhead.
For input with batch size 32 and padding size 128, EnergonAI achieves 46.4$\%$ (1.87:2) latency reduction when scaling from 1 to 2 GPUs, while it is only 82.0$\%$ (5.56:8) latency reduction when scaling from 1 to 8 GPUs.
Although tensor parallelism is not efficient, it is the only parallel pattern that can reduce latency.
Thus, given the cost efficiency, we recommend that fewer devices should be parallelized using tensor parallelism once the latency constraint is satisfied.

\subsection{Technique-1: NBPP}

\begin{figure*}[!t] 
\centering 
\includegraphics[width=0.95\textwidth]{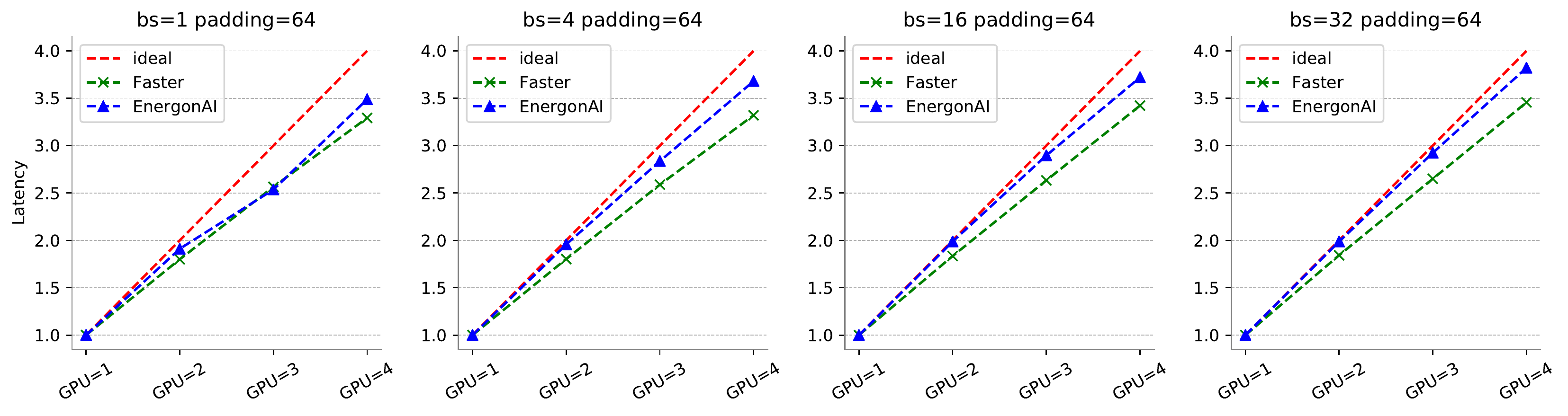} 
\caption{Pipeline Parallelism Scalability of EnergonAI on the partial NVLINK connected server. Here a 12-layer GPT-3 is customized and scales from 1 to 4 GPUs. FasterTransformer is the baseline.}  
\label{fig:latency_pp}
\vspace{-10pt}
\end{figure*}

\begin{figure*}[!t]
\centering 
\includegraphics[width=0.95\textwidth]{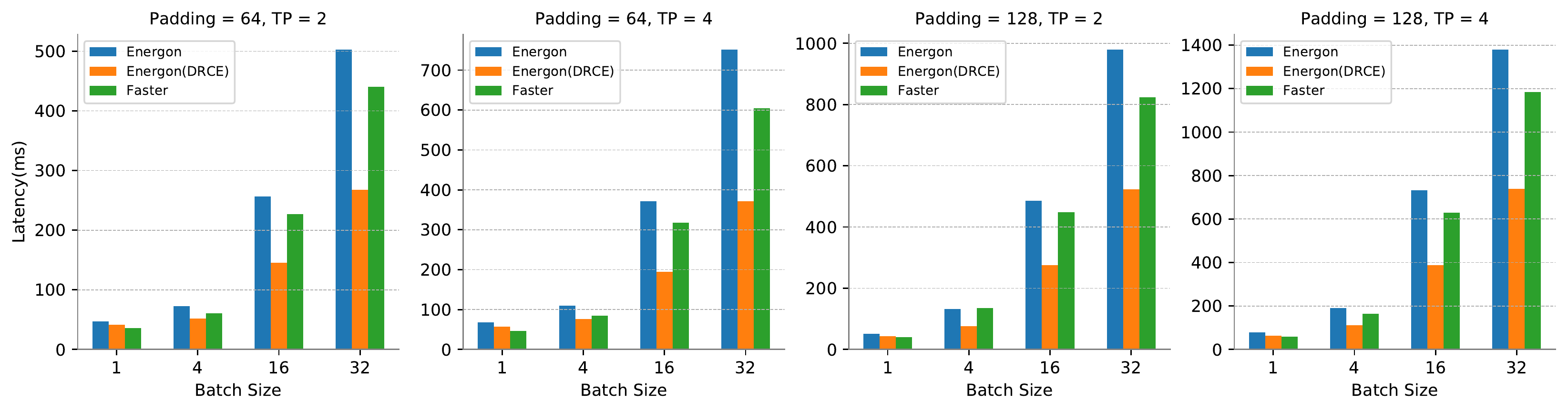} 
\caption{EnergonAI(DRCE) and FasterTransformer on the partial NVLINK connected server. For TP=2, we customize a 24-layer GPT-3 and perform it with 2-device tensor parallelism. For TP=4, we customize a 48-layer GPT-3 and perform it with 4-device tensor parallelism.}  
\label{fig:fastercom}
\vspace{-10pt}
\end{figure*}

\begin{figure*}[!t]
    \centering
    \includegraphics[width=0.95\textwidth]{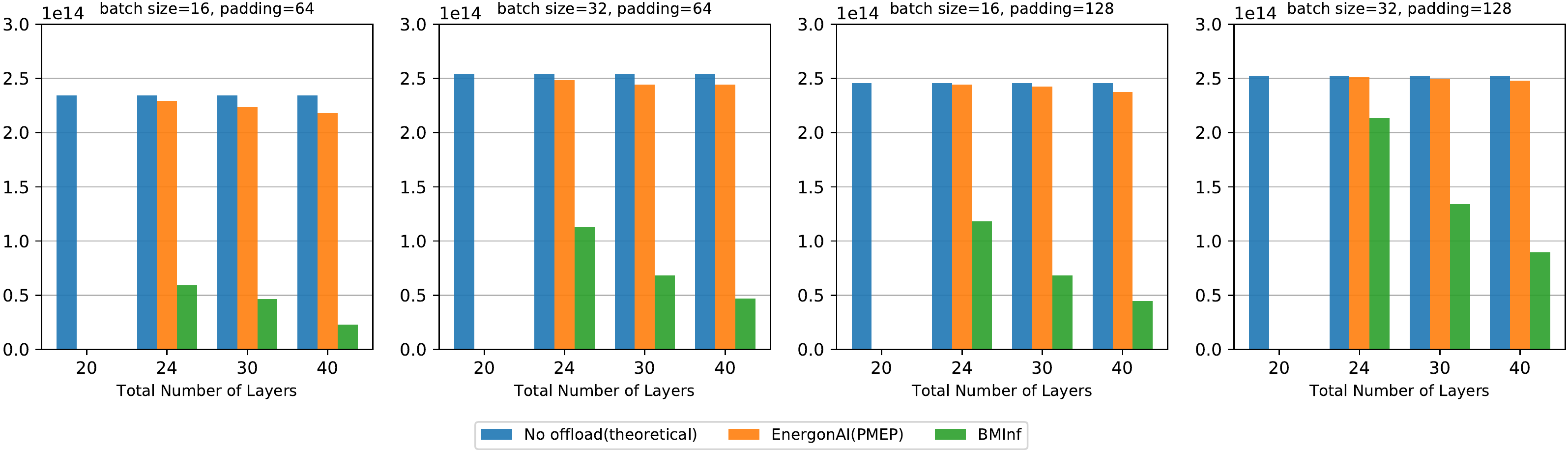}
    \caption{Throughput in TFLOPS(Floating-point operations per second) of different offload targets. The throughput of 24-, 30-, 40- layer GPT-3 without offloading are theoretical numbers from the 20-layer GPT-3.}
    \label{fig:offload_eval}
    \vspace{-10pt}
\end{figure*}

As shown in Figure~\ref{fig:latency_pp}, we evaluate the scalability of pipeline parallelism.
In comparison, we select FasterTransformer as the baseline.
We perform the inference of a large model on 1, 2, 3 and 4 GPUs of the partial NVLINK connected server respectively.
In order to evaluate the strong scalability, the target model should be contained in a single GPU so that we use the 12-layer GPT-3.
Besides, We use 64 as the padding size and 1,4,16,32 as the batch size.

As batch size increases, the scalability of pipeline parallelism becomes better.
In the first graph of Figure~\ref{fig:latency_pp}, the speedup is much lower than the ideal linear scalability when the batch size equals to 1, which is only 3.29$\times$ speedup in FasterTransformer and 3.49$\times$ speed up in EnergonAI when scales to 4 GPUs.
While the speedup increases to 3.45$\times$ and 3.82$\times$ when the batch size becomes 32.
Similarly, as GPU number increases, the relative speedup of pipeline parallelism goes down slightly.
In detail, when the batch size is 32, the speedup ratio of 4 GPUs is 0.93(3.82/4), the ratio of 3 GPUs is 0.96(2.92/3), and the ratio of 2 GPU is 0.99(1.98/2).
In pipeline parallelism, the ratio of the computation time and the non-overlapped communication time decides the final speedup.
Larger batch size has longer execution time on each GPU, and the proportion of communication time drops, leading to the better scalability.
In terms of scaling a fixed model into more devices, a larger pipeline parallelism has more communications, leading to the worse scalability.
Besides, since there is only one embedding module in the top of the transformer model, the slight imbalance problem will be more obvious when increasing the devices of pipeline parallelism.

In comparison with tensor parallelism, pipeline parallelism introduces much less overhead.
As above, our experiments achieve a near-linear acceleration when scaling from 1 to 4 GPUs.
The number of communication is consistent with the number of model layers in tensor parallelism, while only $(\#GPU - 1)$ communications are required in pipeline parallelism.
When distributing 12-layer GPT-3 onto 4 devices, each device only executes 3 layers, which is far from saturating the device memory.
In fact, the number of layers in the complete GPT-3 is 96, which is 8 times more than the 12-layer GPT-3 used in our experiments, meaning a larger proportion of computation time.
In this way, the pipeline parallelism is much easier to achieve the near-linear acceleration.
It is better to use less tensor parallelism once the latency constraint is satisfied and take more pipeline parallelism for enlarging the memory capacity.

Comparing our EnergonAI with FasterTransformer, Figure~\ref{fig:latency_pp} shows that the pipeline parallelism scalability of EnergonAI outperforms that of FasterTransformer in almost all cases.
Our EnergonAI is approximately 10$\%$ better than that of FasterTransformer.
In FasterTransformer, the blocking communication operations $nccl\_send()$ and $nccl\_recv()$ are adopted and lead to more bubbles in the pipeline.
By comparison, because of the hierarchy-controller architecture, our EnergonAI can adopt non-blocking operations for communications between two consecutive devices.

\subsection{Technique-2: DRCE} 

In this section, we perform tensor parallelism on 2 and 4 GPUs of the partial NVLINK connected server to compare EnergonAI(DRCE) with FasterTransformer.
For $TP=2$, we customize a 24-layer GPT-3, and for $TP=4$, we customize a 48-layer GPT-3.
Here we set the valid sequence length half of the padding size in our experiments for DRCE.

Figure~\ref{fig:fastercom} displays that the pure tensor parallelism performance of EnergonAI is about 12$\%$ lower than that of FasterTransformer.
Although the profiling result shows that the general matrix multiplication (GEMM) contributes above 95$\%$ computation time in large models, making the optimization of other kernels less significant, FasterTransformer achieves the advantage from two ways, i.e. selecting the best GEMM kernels and GEMM kernel fusion.
For selecting the best GEMM kernels (not have effect on NVIDIA Ampere architecture), both Pytorch and FasterTransformer process GEMM on GPUs using the cublas library and each GEMM interface in cublas provides multiple different algorithms.
In the warm-up phase, FasterTransformer profiles all algorithms of GEMM kernels and selects the best algorithm.
For the GEMM kernel fusion,
FasterTransformer creates the fused multi-head attention kernel that adds layer-norm, query, key and value GEMMs, and bias adds into a single kernel, which we can also adopt at the cost of programmability.

After adding DRCE, EnergonAI gets great acceleration and outperforms FasterTransformer a lot.
As shown in Figure~\ref{fig:fastercom}, EnergonAI(DRCE) achieves up to 46.8$\%$ latency reduction comparing with the pure EnergonAI and up to 39$\%$ latency reduction comparing with FasterTransformer, showing great effectiveness.
Although the effectiveness of DRCE is related to the distribution of input, the redundant computation amount in our experiment setup should be less than these corpora of GLUE benchmark based on Du et al.~\cite{du2022handling}.
Moving to these experiments with batch size 1, we notice that FasterTransformer ranks first again.
This is because these memory-intensive kernels (other than GEMM kernels) take more proportion in the inference when batch size is very small, making the optimization of FasterTransformer becomes obvious.
However, we think the inference with so small the batch size is not the common case.

Besides the effectiveness of DRCE, Figure~\ref{fig:fastercom} can also display that tensor parallelism is very sensitive to the interconnection performance.
In other words, it is hard for tensor parallelism to scale large, especially when interconnection is not very high.
Doubling the number of GPUs (from TP=2 to TP=4) and the workload (from 12 layers to 24 layers), the latency increases about 1.4$\times$.
Since only every two GPUs are connected using NVLINK and other interconnections are PCIe, the interconnection performance reduces a lot when increasing from 2 to 4 GPUs, resulting in massive performance degradation.

\subsection{Technique-3: PMEP}

For PMEP, we evaluate its performance on two NVIDIA A100 GPU connected with NVLINK of the partial connected server.
The scenario we choose here is doing GPT inference with limited computing device memory. 
When using the GPT-3 configuration, our 80 GB memory can hold at most 20 layers. 
Suppose we have four specific models with 20, 24, 30 and 40 layers separately and only one GPU to do computation, which means we would need extra memory for the latter three models.
Here we compare our PMEP with the CPU offloading method in BMInf~\cite{hanetal2022bminf}.
Basically, we hold 20 layers on GPU 0 (the local device) while the remaining layers in each model are offloaded to GPU1 (the peer GPU) or CPU.
We take 32 and 64 as the batch size, and 64 and 128 as the padding size.
Simultaneously, we simulate the practical workloads on GPU1 (the peer GPU) by executing ResNet50~\cite{he2015resnet} with TensorRT (batch size is 128), which takes about 3.5GB memory.

Our strategy of offloading is decided before the inference starts and these offloading layers are distributed evenly among those to be held on device. 
Taking the 24-layer GPT-3 for example, layers No.5, 11, 17, and 23 are offloaded. 
To achieve the best communication and computation overlap, we prefetch the next off-device layer immediately the execution of the previous off-device layer is finished and offload a off-device layer immediately after the computation's done.
The throughput results are shown in figure \ref{fig:offload_eval}. 
The y axis represents the FLOPS(Floating-point operations per second), which we calculate with the parameters of the models. 
As one GPU can not hold the latter three models, we use the throughput of the 20-layer GPT-3 to calculate the theoretical numbers for them. 

For the EnergonAI(PMEP), we notice that the performance of the local GPU is only affected a little by offloading in almost all cases.
For the case with batch size of 32 and padding size of 64, the throughput of the local GPU reduces by 2.3$\%$, 3.9$\%$, and 3.9$\%$, which are quite acceptable since it enables the inference of models with 1.2$\times$, 1.5$\times$, and 2$\times$ layers. 
Simultaneously, in our experiments, the workloads on the peer GPU can also keep its performance.
By contrast, the CPU offloading of BMInf is much different. 
There are obvious throughput reduction in all experiments.
For the same case above, the throughput reduces by 55$\%$, 73$\%$, 81$\%$ for three models.
This is because the communication overhead is relatively small and can be almost completely overlapped with the computation in the EnergonAI(PMEP), while the time of communication exceeds that of computation in the BMInf.

In addition to the throughput, EnergonAI(PMEP) is more beneficial for the latency.
As shown in the Figure~\ref{fig:offload_eval}, EnergonAI(PMEP) shows greater advantage than the CPU offloading in BMInf when batch size and padding size is smaller.
As batch size or padding size grow, the increased computation time can better overlap with the communication time and lead to better throughput for the CPU offloading method, while it will damage the latency of each sample.
In comparison, EnergonAI(PMEP) can keep the throughput with smaller batch size and padding size, achieving better latency.



\section{Conclusion}

EnergonAI targets at the efficient inference of 10-100 billion parameter transformer models.
It adopts a hierarchy-controller system architecture to coordinate multiple devices and efficiently support different parallel strategies, e.g. tensor parallelism and pipeline parallelism.
Additionally, EnergonAI includes three novel techniques, i.e. distributed redundant computation elimination, non-blocking pipeline parallelism, and peer memory pool, to further improve the latency and throughput, and relieve the memory wall problem.
In most cases of tensor parallelism, EnergonAI can achieve comparable performance to FasterTransformer in fixed-length experiments, and superior performance to FasterTransformer in variable-length experiments.
EnergonAI exhibits better scalability in pipeline parallelism compared with FasterTransformer.
Besides, EnergonAI can infer larger models in a single GPU by using a larger heterogeneous memory space without obvious performance degradation.


\newpage
\bibliographystyle{unsrt}
\bibliography{tex/sample-base}


\end{document}